\documentclass[10pt,journal,compsoc]{IEEEtran}

\usepackage{amsmath}
\usepackage{amssymb,subfigure,cases}
\usepackage{color,hyperref}
\usepackage{algpseudocode}
\usepackage{booktabs}

\usepackage{amsmath}
\usepackage{subfigure}
\usepackage{caption}
\usepackage{algorithm}
\usepackage{algpseudocode}
\usepackage{setspace}

\usepackage{soul}
\usepackage{pifont}

\usepackage{times}
\usepackage{epsfig}
\usepackage{graphicx}
\usepackage{amssymb}
\usepackage{multirow}
\usepackage{subfigure}
\usepackage{pifont}
\usepackage{color,xcolor,colortbl}
\usepackage{array}
\usepackage{url}
\usepackage{CJK}
\usepackage{makecell}

\definecolor{mygray}{gray}{.9}

\newcommand{\xmark}{\ding{55}}
\newcommand{\rmark}{\ding{52}}

\newcolumntype{I}{!{\vrule width 1.2pt}}
\newlength\savedwidth
\newcommand\whline{\noalign{\global\savedwidth\arrayrulewidth
		\global\arrayrulewidth 1.25pt}%
	\hline
	\noalign{\global\arrayrulewidth\savedwidth}}

\ifCLASSOPTIONcompsoc
  \usepackage[nocompress]{cite}
\else
  \usepackage{cite}
\fi

\ifCLASSINFOpdf
\else
 \fi

\begin{document}
	
\setul{}{1.0pt}
%
\title{Learning Independent Instance Maps for Crowd Localization}

\author{
     Junyu Gao, \IEEEmembership{Member,~IEEE,} Tao Han, Qi~Wang, \IEEEmembership{Senior~Member,~IEEE,} Yuan~Yuan, \IEEEmembership{Senior~Member,~IEEE,}   Xuelong~Li, \IEEEmembership{Fellow,~IEEE} 
\IEEEcompsocitemizethanks{
\IEEEcompsocthanksitem 
The authors are with the School of Artificial Intelligence, Optics and Electronics (iOPEN), Northwestern Polytechnical University, Xi'an 710072, Shaanxi, China. E-mails: gjy3035@gmail.com, hantao10200@mail.nwpu.edu.cn, crabwq@gmail.com, y.yuan1.ieee@gmail.com, li@nwpu.edu.cn. 

\textit{Xuelong Li is the corresponding author.}.
}}

\markboth{IEEE TRANSACTIONS ON PATTERN ANALYSIS AND MACHINE INTELLIGENCE,~Vol.~XXX, No.~XXX, XXX~XXX}%
{Shell \MakeLowercase{\textit{et al.}}: Bare Advanced Demo of IEEEtran.cls for IEEE Computer Society Journals}

\IEEEtitleabstractindextext{%
\begin{abstract}

Accurately locating each head's position in the crowd scenes is a crucial task in the field of crowd analysis. However, traditional density-based methods only predict coarse prediction, and segmentation/detection-based methods cannot handle extremely dense scenes and large-range scale-variations crowds. To this end, we propose an end-to-end and straightforward framework for crowd localization, named Independent Instance Map segmentation (IIM). Different from density maps and boxes regression, each instance in IIM is non-overlapped. By segmenting crowds into independent connected components, the positions and the crowd counts (the centers and the number of components, respectively) are obtained. Furthermore, to improve the segmentation quality for different density regions, we present a differentiable Binarization Module (BM) to output structured instance maps. BM brings two advantages into localization models: 1) adaptively learn a threshold map for different images to detect each instance more accurately; 2) directly train the model using loss on binary predictions and labels. Extensive experiments verify the proposed method is effective and outperforms the-state-of-the-art methods on the five popular crowd datasets. Significantly, IIM improves F1-measure by 9.0\% and ranks first on the NWPU-Crowd Localization benchmark. The source code and pre-trained models are released at \url{https://github.com/taohan10200/IIM}.
\end{abstract}

}

\maketitle

\IEEEpeerreviewmaketitle

\section{Introduction}
\setulcolor{red}
\label{Sec:introduction}
Crowd localization is a sub-task in crowd analysis, of which purpose is to predict each instance's position. Compared with image-level crowd counting \cite{li2018csrnet,2019bayesian}, instance is a basic unit in localization task, which can provide more accurate results to serve high-level crowd analysis tasks \cite{kang2014fully,li2017multiview}. However, due to the lack of box-level data annotation, localization development is slower than counting tasks. With the release of box-level annotated datasets (NWPU-Crowd \cite{gao2020nwpu} and JHU-CROWD \cite{sindagi2020jhu}), this task will attract the attention of more researchers.

\begin{figure}[t]
	\centering
	\includegraphics[width=0.45\textwidth]{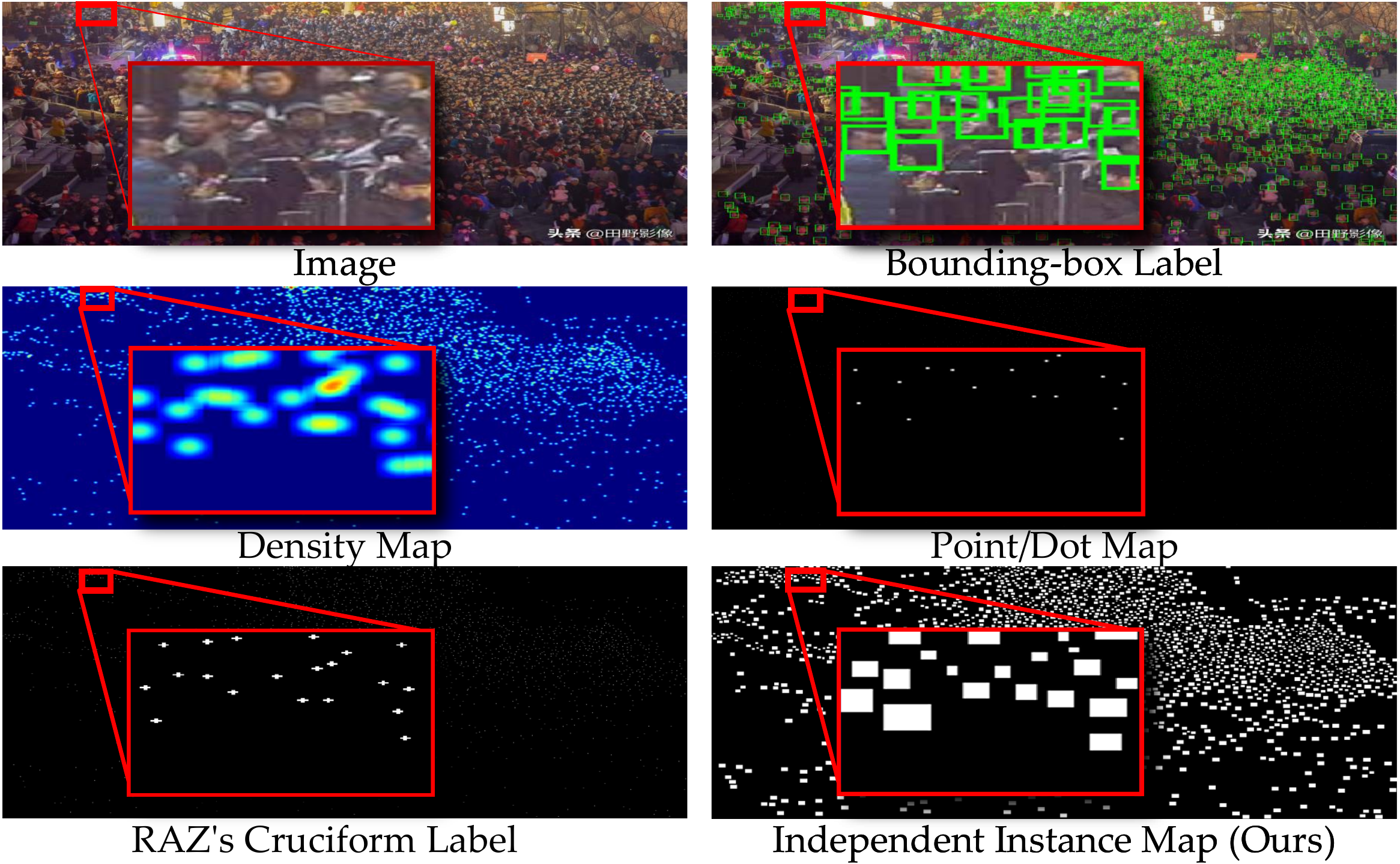}
	\caption{The comparison of four types of mainstream groundtruth for crowd localization and ours. The proposed independent instance map is non-overlapped and scale-aware, which is more suitable for dense object localization.}
	\label{fig:intro}
	\vspace{-0.5cm}
\end{figure}

At present, there are three types of methods that focus on crowd localization. Specifically, 1) detection-based methods \cite{stewart2016end,liu2018decidenet} directly use some object detection models (Faster RCNN \cite{ren2015faster}, TinyFaces \cite{hu2017finding}, \emph{etc.}) to predict the head's location. These methods are friendly to the large-scale head but error-prone in the extremely congested crowds. 2) Heuristic algorithms find head locations in the density or segmentation-confidence maps output by the model. Among them, some works \cite{idrees2018composition,liu2019recurrent} locate head position from a maximum or minimum point in a local region ($3 \times 3$, \emph{etc.}), which causes multiple predictions for large-scale head samples or dense regions. Besides, it is hard to get the accurate position in the smooth heat map. Others \cite{arteta2016counting,abousamra2020localization} segment the confidence map to individual blocks by a fixed threshold and then locate heads by detecting the connected components. However, a single threshold has difficulty in segmenting the small objects with blurry and dense attributions. 3) Point-supervision methods \cite{laradji2018blobs,liu2019point,sam2020locate} usually directly use point-level labels or generate pseudo box label to learn to locate the head position. However, these labels do not reflect head sizes well. Moreover, all the above algorithms suffer from large-range scale variations.

This paper follows the heuristics branch and utilize connected component segmentation \cite{arteta2016counting,abousamra2020localization} \ul{for crowd localization}. Specifically, in crowd scenes, each head corresponds to an independent connected component, of which size is related to the head region. The whole GT map is named as ``Independent Instance Map (IIM)''. Fig. \ref{fig:intro} shows four traditional labels and the proposed IIM. From it, IIM simultaneously shows the better independence and scale awareness than others. To reduce the aforementioned problems brought by others and locate the head more accurately, a differentiable Binarization Layer is proposed to output independent instance maps. Different from the Differentiable Binarization (DB) \cite{liao2020real}, the proposed binarization layer does not require additional supervision information and learns a threshold to transform the confidence map to binary segmentation map. Compared with the localization methods \cite{idrees2018composition,liu2019recurrent,gao2019domain,xu2019autoscale,liang2021reciprocal,abousamra2020localization} that extrac the head positions with a non-differentiable heuristic scheme, the proposed binarization layer is learnable, and allows the network to be optimized with a hard loss between binary prediction and labels. Therefore, it can improve the edge segmentation performance for tiny objects, resulting to a better localization performance.

Furthermore, we find that the networks have different confidence distribution for different-scale objects. In general, the confidences for the large-scale and small-scale heads are less than the mederate-size instances. Hence, to capture more objects from the different confidence regions, we introduce the point-to-point thresholds, which are produced by the designed Pixel-level Binarization Module (PBM). It is aware of the head spatial distribution and determines different thresholds for each pixel binarization process. Thus, PBM can output elaborate instance maps to achieve more accurate localization results.

In a summary, this paper has three contributions:
\begin{enumerate}
	\item[1)] Utilize connected components for crowd localization. Based on it, we propose a simple and effective end-to-end Independent Instance Maps segmentation (IIM) to locate head positions in crowd scenes.
	\item[2)] Present a differentiable Binarization Module to output IIM for accurate object localization. Its pixel-level operation can learn adaptive thresholds for those heads with large-range scale variation.
	\item[3)] The proposed method outperforms the state-of-the-art localization models even in the case of dot labeling. 
\end{enumerate}

\section{Related Works}
\subsection{Crowd Counting}
Since crowd localization and counting are closely related, it is necessary to briefly review recent counting works. With the development of deep learning, CNN-based methods \cite{zhang2016single,onoro2016towards,wang2020pixel,bai2020adaptive} show its powerful capacity of feature extraction than hand-crafted features models \cite{idrees2013multi}. Some methods \cite{shi2018crowd,liu2020adaptive} work on network architectures or specific modules to regress pixel-wise or patch-wise density maps. In addition to density-map supervision, some methods \cite{ma2019bayesian,wang2020distribution} directly exploit point-level labels to supervise counting models. Unfortunately, these methods only predict image-level counts or coarse local density. The position of each head is hard to obtain accurately.

\subsection{Crowd Localization}

To locate the head position of each head in the crowd scenes, some researchers focus on crowd localization, consisting of three types: (a) detection-based, (b) heuristic, and (c) point-supervision methods. Table \ref{Table:related} illustrates the features of these methods and the proposed IIM.

\begin{table}[htbp]	
	\centering
	\caption{Comparison of three typical methods and the proposed IIM (ours). The (a), (b), (c) and (d) denote detection-based, heuristic, point-supervision and independent instance map segmentation.  }
	\scriptsize
	\begin{tabular}{c|c|c|c|c}
		\whline
		\multirow{2}{*}{Method} & \multirow{2}{*}{Type} &\multicolumn{2}{c|}{Labels} &Learnable  \\
		\cline{3-4}
		& & Non-overlapped &Scale-aware &  Localization	\\
		\whline
		TinyFaces \cite{hu2017finding} & (a)  &\xmark &\rmark  & \rmark	\\
		\hline
		Idrees \emph{et al.} \cite{idrees2013multi} &(b)  &\xmark &\xmark  & \xmark	\\
		\hline
		LSC-CNN \cite{sam2020locate} &(c) &\rmark &\xmark  & \rmark	\\
		\hline
		IIM (ours) & (d)  &\rmark &\rmark  & \rmark \\
		\whline			
	\end{tabular}\label{Table:related}
\end{table}

\textbf{Detection-based Crowd Localization\,\,\,\,}
Many methods detect the head with a bounding box to locate the position. Ishii \emph{et al.} \cite{ishii2004face} propose a head detection method using hand-crafted features (four directional features, FDP) for a real-time surveillance system.
Due to the limitation of hand-crafted features, some methods utilize CNN to improve complicated scenes' performance. Vu \emph{et al.} \cite{vu2015context} present a context-aware head detection algorithm based on R-CNN detector \cite{girshick2014rich} in natural scenes. Stewart and Andriluka \cite{stewart2016end} design a recurrent LSTM layer for sequence generation and a end-to-end detector (OverFeat \cite{sermanet2013overfeat}) to output a set of distinct detection results. However, general object detectors (RCNN, OverFeat) work poorly in dense crowds. Therefore, some researchers design specific networks for tiny objects. Hu and Ramanan \cite{hu2017finding} propose a tiny object detection framework by analyzing the impacts of image resolution, head scale, and context, significantly improving the performance in dense crowds. After this, there are many works \cite{bai2018finding,2019PyramidBox,Li2020DSFD} focus on this field. However, they are cannot works well in extremely congested scenes.

\textbf{Heuristic Crowd Localization\,\,\,\,}
Benefiting from high-resolution, fine density map and segmentation map, some methods try to find head center from them. Idrees \emph{et al.} \cite{idrees2018composition} and Liu \emph{et al.} \cite{liu2019recurrent} propose a post-processing method by finding the peak point in a local region ($3 \times 3$ pixels) as the final head position. Gao \emph{et al.} \cite{gao2019domain} present a Gaussian-prior Reconstruction algorithm to recover the center of $15 \times 15$ Gaussian Kernel from predicted density maps. Xu \emph{et al.} \cite{xu2019autoscale} and Liang \emph{et al.} \cite{liang2021reciprocal} propose to regress distance map and then extract the head positions from distance map. These algorithms exploit fixed regions, they may output multiple predictions for one large-scale head. Hence, Abousamra \emph{et al.} \cite{abousamra2020localization} propose to segment the head regions and introduce a persistence loss to learn topological constraint for dots.

\begin{figure*}[t]
	\centering
	\includegraphics[width=0.95\textwidth]{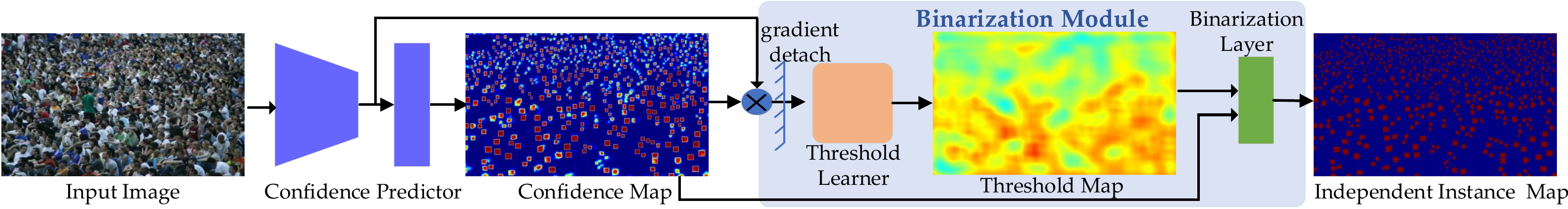}
	\caption{Crowd localization framework based on Independent Instance Maps segmentation (IIM). It first makes confidence predictions for head areas in the crowd scene. The image or pixel-level binarization module segments the confidence map to an independent instance map. During inference, the threshold module predicts the thresholds online for each confidence map. Finally, by detecting 4-connected components, the box and the center of each independent instance region are obtained.}
	\label{fig:framework}
		\vspace{-0.5cm}
\end{figure*}

\textbf{Point-supervision Crowd Localization\,\,\,\,}
As the box-level annotations are scarce, many methods directly exploit point labels to supervise the model for crowd localization. Laradji \emph{et al.} \cite{laradji2018blobs} combine four types of loss functions to push the model segment a single blob for each object. Liu \emph{et al.} \cite{liu2019point} propose a curriculum learning strategy to generate pseudo box-level labels from point annotations. Sam \emph{et al.} \cite{sam2020locate} propose a tailor-made dense object detection method only using point-level annotations, which predicts position, size for each head simultaneously.  Wang \emph{et al.} \cite{wang2021self} propose a self-training strategy to estimate the centers and sizes of crowded objects, which initializes pseudo object sizes from the point-level labels.

\section{Methodology}
\textbf{Overview. }
This paper aims to establish a crowd localization framework based on Independent Instance Maps segmentation (IIM). As shown in Fig. \ref{fig:framework}, the first step is to predict the confidence of the crowd area. Then, a binarization module is applied to confidence maps to output binary map. Finally, the connected components in the segmentation map are detected to output the boxes that include the head centers. Section \ref{Sec:Threshold_Layer} and \ref{Sec:ThresholdModule} describe the core component binarization layer and module. Section \ref{CLF} shows the proposed Crowd Localization Framework.

\subsection{Binarization Layer}
\label{Sec:Threshold_Layer}

\subsubsection{Problem setting}

\begin{figure}[tbp]
	\centering
	\includegraphics[width=0.45\textwidth]{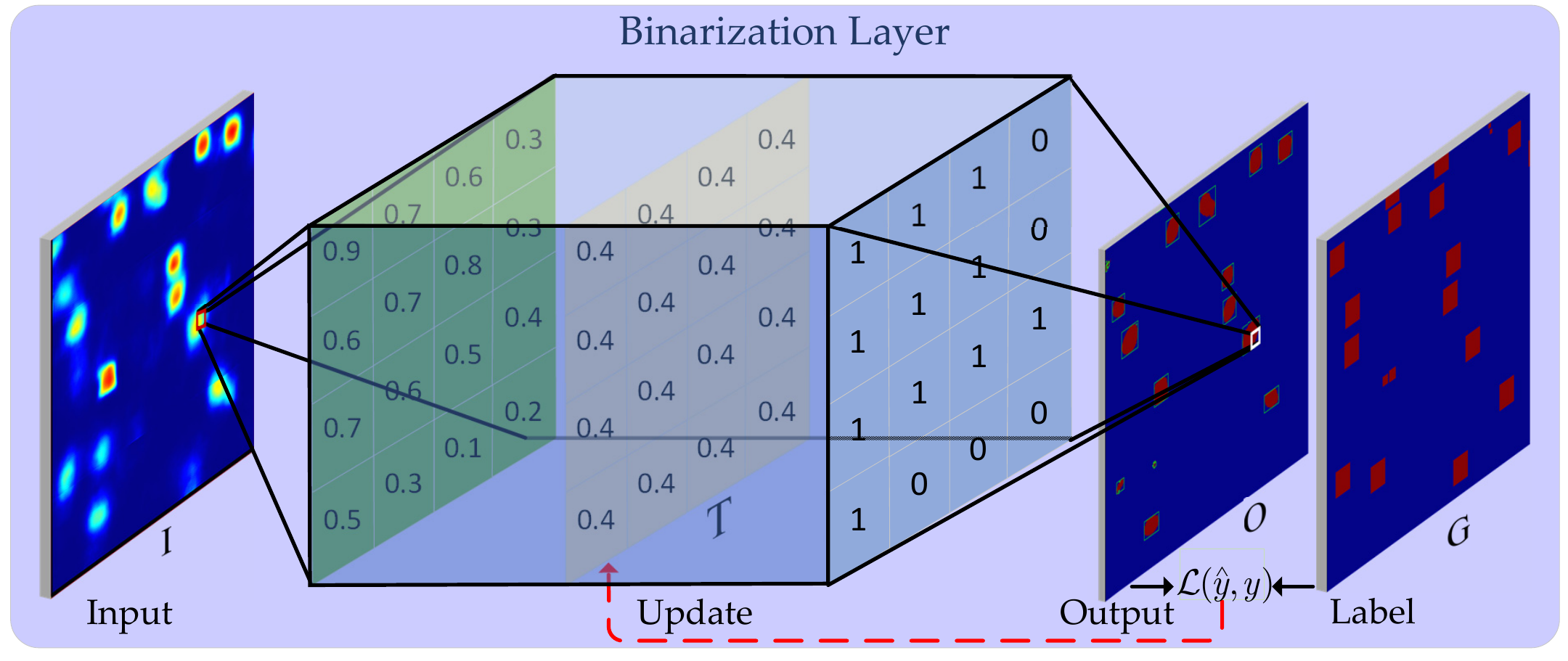}
	\caption{The workflow of the binarization layer. A learnable threshold is adopted to deal with some particular vision tasks that need precise segmentation.}
	\label{fig:binar_global}
		\vspace{-0.5cm}
\end{figure}

As shown in Fig. \ref{fig:binar_global}, a binarization layer is proposed under the following setting: Particularly, there is an input $I\in \mathbb{R}^{H \times W}$, and the range for each pixel is $[0,1]$. Correspondingly, there is a target binary image $G\in \mathbb{R}^{H \times W}$ where the value is $0$ or $1$. The goal of the binarization layer is to learn a threshold $T$ to segment the input image, so that its output image $O$ is as close as possible to the target image $G$. Generally, given an input sequence $\{I_{n}\}_{n=1}^{N}$ and a target sequence $\{G_{n}\}_{n=1}^{N}$, we expect to achieve optimal segmentation on all data with tailor-made thresholds for different images.


\subsubsection{Theoretical Analysis}

\textbf{Forward Propagation.} For a point $I(i,j)$ in the input $I$ ($i$, $j$ is the index of two axes), the forward process of the proposed binarization layer $\mathcal{B}$ is defined as follows:

\begin{equation}
\small
O(i,j)=\operatorname{\mathcal{B}}(I,T)=\left\{\begin{array}{ll}
1, & \text { if } I(i,j) \geq T \\
0, & \text { otherwise }
\end{array},\right.
\label{Eq:1}
\end{equation}
where threshold $T$ is a parameter to be learned. This forward process is non-differentiable because it is a comparsion operation. Thus, it is impossible to compute the gradient of $T$ automatically. Here we define a similar mathematical to simulate the relationship between the binary output $\mathcal{O}$ and $T$, of which the output only has relationship with $T$.

Assuming $I\in \mathbb{R}^{H \times W}$ satisfies the two conditions: 1) $H \times W \rightarrow \infty$ , and 2) $I(i,j) \sim U(0,1)$, the uniform distribution. Then, from the perspective of probability and statistics, the cumulative distribution function $F_{I(i,j)}(T)$ has the relationship with $T$ as follows:

Assuming $I\in \mathbb{R}^{H \times W}$ satisfies the two conditions: 1) $H \times W \rightarrow \infty$ , and 2) $I(i,j) \sim U(0,1)$, the uniform distribution. Then, from the perspective of probability and statistics, the cumulative distribution function $F_{I(i,j)}(T)$ has the relationship with $T$ as follows:

Assuming $I\in \mathbb{R}^{H \times W}$ satisfies the two conditions: 1) $H \times W \rightarrow \infty$ , and 2) $I(i,j) \sim U(0,1)$, the uniform distribution. Then, from the perspective of probability and statistics, the cumulative distribution function $F_{I(i,j)}(T)$ has the relationship with $T$ as follows:
\begin{equation}
\small
F_{I(i,j)}(T)=P(I(i,j) \leq T)=\left\{\begin{array}{ll}
0, & \text { if }  T <0  \\
T, & \text { if } 0 \leq T <1 \\
1, & \text { if } T \ge 1
\end{array},\right.
\label{Eq:2}
\end{equation}
where $P(I(i,j)\leq T)$ represents the probability that a pixel value in $I$  is less than $T$.

To learn the threshold $T$, we sum the all pixels in output image $\mathcal{O}$ to a scalar in order to calculate the loss for back propagation. In practic, let $ \hat{y} = \frac{1}{W\times H}\sum\limits_{i \in H}\sum\limits_{j \in W} O(i,j)$, then the range of $\hat{y}$ is $[0,1]$. With the conditions in Eq. \ref{Eq:2}, $\hat{y}$ has the following relationship with the threshold $T$:
\begin{equation}
\small
\hat{y}=\left\{\begin{array}{ll}
1, & \text { if }  T <0  \\
1 - T, & \text { if } 0 \leq T \leq1 \\
0, & \text { if } T > 1
\end{array}.\right.
\label{Eq:3}
\end{equation}

\textbf{Back Propagation.}  Eq. \ref{Eq:3} reval that the non-differential binarization is relaxed as a linear operation. The prediction $\hat{y}$ and the threshold value $T$ under such conditions is a linear operation. This relaxation only takes place on the back-foward stage because we need to provide gradient for $T$. Here $\hat{y}$ is only depends on $T$ and the derivative of $T$ is a constant $-1$. To make the target $G$ have the same form as $O$, G is defined as $ y = \frac{1}{W\times H}\sum\limits_{i \in H}\sum\limits_{j \in W} G(i,j)$. $T$ can be updated with the learning rate $\alpha$ as in Eq. \ref{Eq:4},
\begin{equation}
\small
T=: T-\alpha \nabla_{T} \mathcal{L}\left(\hat{y}, y\right)=T+\alpha\frac{\partial \mathcal{L}}{\partial \hat{y}},
\label{Eq:4}
\end{equation}
where $\mathcal{L}(\hat{y}, y)$ is the loss function to measure the distance between the output $O$ and the target $G$. It can be a simple L1 Loss or other distance measure functions.

\subsection{Binarization Module}
\label{Sec:ThresholdModule}
In section \ref{Sec:Threshold_Layer}, the binarization layer can segment the input continuous heat map to a binary map with a learnable threshold. To make the the threshold update according to the image content, the Binarization Module (BM) is further proposed, which consists of a threshold encoder and a binarization layer. The former encodes the image's feature map and generates a single value or a map. The latter exploits this value/map to binarize the confidence map and outputs the instance map.

\begin{figure}[t]
	\centering
	\includegraphics[width=0.45\textwidth]{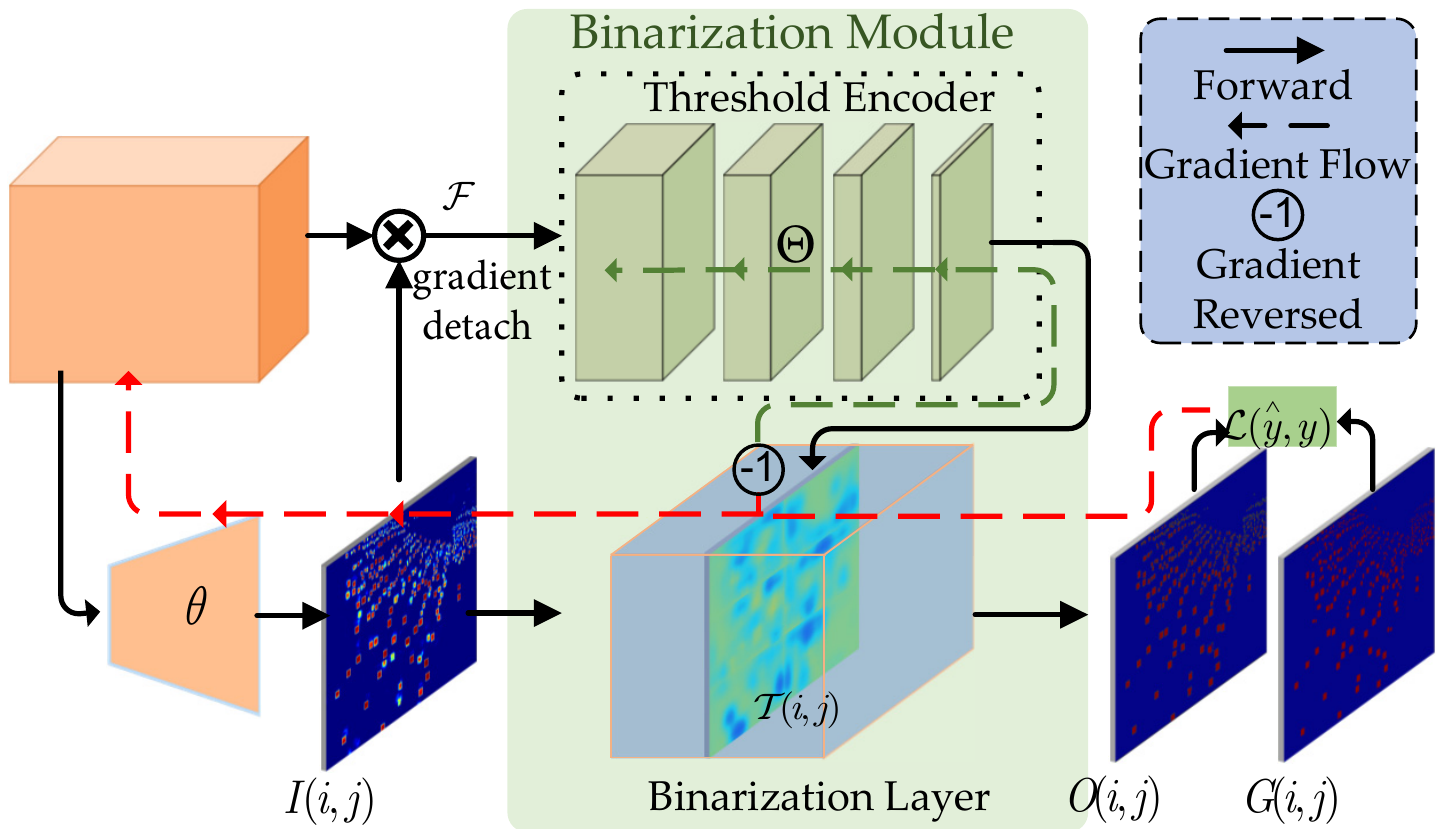}
	\caption{The flowchart of embedding Binarization Module into the networks. It is designed to give proper thresholds according to the features, making a better segmentation.}
	\label{fig:binar_pixel}
\end{figure}

Fig. \ref{fig:binar_pixel} shows the architecture for a general BM. In this module, a threshold prediction is output for the confidence map $I$ according to the feature map $\mathcal{F} \in \mathbb{R}^{W \times H \times C}$ (which is usually extracted from the backbone). Considering a convolutional neural network as threshold encoder, the feature $\mathcal{F}$ can be mapped to the threshold $\mathcal{T}$ by the threshold encoder with the parameters $\Theta$:
\begin{equation}
\small
\mathcal{T}=\Phi( \mathcal{F};\Theta),
\label{Eq:5}
\end{equation}
where $\mathcal{T}\in (0,1)$ can be a scalar or a matrix with the size of $I$. For a scalar $\mathcal{T}$ (namely, $i,j=1$), the relationship between the output and  $\mathcal{T}$ is shown in Eq. \ref{Eq:3}. For $\mathcal{T}$ at the pixel level, the forward propagation process and the cumulative distribution are similar to Eq. \ref{Eq:1} and Eq. \ref{Eq:2}, respectively. Moreover, the $\hat{y}$ has the following relationship with $\mathcal{T}(i,j)$:
\begin{equation}
\small
\hat{y} = 1 - \frac{1}{W \times H}\sum\limits_{i \in H}\sum\limits_{j \in W} \mathcal{T}(i,j).
\label{Eq:6}
\end{equation}

\setulcolor{red}
As shown in Fig. \ref{fig:binar_pixel}, there are two inputs for the binarization module: the under segmented image $I$ and the threshold $\mathcal{T}$. For the threshold encoder, the derivative at every pixel of $\mathcal{T}$ is -1. It means that the gradients will be reversed when it flows to the threshold encoder by threshold $\mathcal{T}$. So, the parameters $\Theta$ in threshold learner can be optimized by gradient descent as follows:
\begin{equation}
\small
\Theta := \Theta+\beta \nabla_{\Theta} \mathcal{L}\left(\hat{y}, y\right),
\label{Eq:7}
\end{equation}
where $\beta$ is learning rate of threshold encoder. Besides to optimize the threshold encoder, a hard loss can be calculated with binary prediction and labels, and make a back propagation from input $I$ to the confidence map's prediction network (e.g. a confidence predictor). Unlike the threshold encoder, the gradients of the confidence predictor have the same symbol as the loss $\mathcal{L}(\hat{y}, y)$. Assuming that the input $I$ is output by the confidence predictor with the parameters $\theta$, $\theta$ is updated with the learning rate $\gamma$ as:
\begin{equation}
\small
\theta := \theta- \gamma \nabla_{\theta} \mathcal{L}\left(\hat{y}, y\right).
\label{Eq:8}
\end{equation}

Eq. \ref{Eq:7} and \ref{Eq:8} reveal that the threshold encoder and the confidence predictor play an antagonistic role. The confidence encoder wants to make the object regions with higher confidence and background areas with lower confidence. The threshold learner strives to make the object regions have low thresholds, and the background areas have high thresholds. By this way, the background noise can be filtered as much as possible, and the low confidence foreground, like tiny-scale and large-scale heads, can be reserved for localization. As these two tasks are antagonistic, we add the gradient detach operation between the backbone and the threshold learner, as illustrated in the Fig. \ref{fig:framework} and Fig. \ref{fig:binar_pixel} . The visualization analysis are provided in the experimental section to discuss the antagonistic role.


\subsection{Crowd Localization Framework}

\label{CLF}

\subsubsection{Confidence Predictor}

Since we hope to achieve crowd localization by detecting connected components, a high-resolution representation backbone is particularly crucial for this pixel-level visual task. It can help us eliminate some adverse effects caused by scale diversity and occlusion. Therefore, there are two popular networks are exploited as the Confidence Predictor (CP): 1) VGG-16 + FPN, which uses VGG-16 \cite{simonyan2014very} as the backbone and exploits Feature Pyramid Networks \cite{lin2017feature} to encode multi-scale features; 2) HRNet-W48 \cite{wang2020deep}, a high-resolution networks with powerful capacity of feature representations in visual recognition. The more detailed network configurations are available in the supplementary materials.

\subsubsection{Threshold Encoder}
\setulcolor{red}
Section \ref{Sec:ThresholdModule} describes how to embed the binarization layer into the network. Here, two schemes are proposed to binarize the outputs of CP: Image-level and Pixel-level Binarization Module (``IBM'' and ``PBM'' for short). IBM and PBM binarizes the prediction with a single value and a map, respectively. To learn the threshold from image contents, a Threshold Encoder (TE) is designed, of which input is multiplied by the original feature map and the confidence map: $\mathcal{F} := I \odot \mathcal{F}$. The hadamard product is performed to filter the background noise with the predicted confidence map. Here is the details of the TE.

\textbf{IBM:} In IBM, a $1 \times 1$ convolutional layer and a Global Average Pooling (GAP) is applied to output a single value as the learnable threshold for binarization layer.

\textbf{PBM:} The above IBM learns a specific value for an input crowd scene. However, the confidence distributions of heads with different scales are very different. The main reasons are: 1) there is data bias in the dataset, namely a few small-scale and large-scale heads, which causes that the confidence is generally lower than the moderate-scale head. 2) It is difficult for the network to learn discriminative features between some hard samples and backgrounds. Thus, a TE in PBM is presented to produce pixel-level threshold maps, which consists of four convolution layers with PReLU and two large-kernel average pooling with a stride of $1$. The configuration are listed as: Conv: 3$\times$3, PReLU; Conv: 3$\times$3, PReLU; Conv: 3$\times$3, PReLU; Avgpool: 15 $\times$15; Conv:1$\times$1,  Avgpool: 15$\times$15, Sigmoid. In TE of PBM, to cover a large spatial receptive field and save the memory, the input feature is resized to the 1/8 original size. In addition, after the last two convolutional layers, 15 $\times$ 15 average pooling with a stride of 1 is applied to smooth the outputs.

In the experiments, TE may produce some very low (less than 0.1) or high values (close to 1) in the threshold map, which makes the network fluctuate easily. In addition, high threshold results in many hollows in the head region. Therefore, we attempt to constrain the output range of Sigmoid activation function. To be specific, the compressed Sigmoid is proposed, which is formulated as:

\begin{equation}
\small
f(x)=\frac{1}{2+e^{-x}}+0.2.
\label{Eq:9}
\end{equation}

From Eq. \ref{Eq:9}, the output of Compressed Sigmoid is constrained in $(0.2, 0.7)$. The improvement of Sigmoid brings more stable training and better performance.

After getting the threshold map, it will be fed into the binarization layer together with the confidence map. Then the binarization layer produces the segmentation map. Finally, by detecting connected components, the boxes  and the centers of independent instances are obtained.

\subsubsection{Loss Functions}
Two loss functions are involved in this framework: 1) The confidence map learning is a regression problem, MSE loss function trains it well according to our experiments. 2) The threshold map learning also a regression problem and the threshold rang is (0.2, 0.7). Therefor, the L1 loss function is applied to $\mathcal{L}(\hat{y},y)$ in Eq. \ref{Eq:4} for training the threshold learner. Other than that, this objective function provide gradients for confidence predictor, as shown in Eq. \ref{Eq:8}.Note that we don't use another L2 loss for training threshold because we consider the threshold learning need to converge fast than the confidence prediction.

\section{Experiments}

\begin{table*}[htbp]
	\centering
	\scriptsize
	\caption{The leaderboard of NWPU-Crowd Localization (\emph{test set}). $A0 \sim A4$ respectively means that the head area is in  $[10^0,10^1]$, $(10^1,10^2]$,  $(10^2,10^3]$,  $(10^3,10^4]$,  $(10^4,10^5]$, and $>10^5$. ``*'' represents anonymous submission on the benchmark. The time stamp is Mar. 10, 2021, which is recorded in \url{https://www.crowdbenchmark.com/historyresultl.html}.}
	\begin{tabular}{cIcIcIc|cIc|c}
		\whline
		\multirow{2}{*}{Method}	&\multirow{2}{*}{Backbone} &\multirow{2}{*}{Label} &\multicolumn{2}{cI}{Overall ($\sigma_l$)}  &\multicolumn{2}{c}{Box Level (only Rec under $\sigma_l$) (\%)}  \\
		\cline{4-7}
		& & & \textbf{F1-m}/Pre/Rec (\%)  &MAE/MSE/NAE & Avg. &$A0 \sim A5$\\
		\whline
		Faster RCNN \cite{ren2015faster}  & ResNet-101& Box & 6.7/\textbf{95.8}/3.5 &414.2/1063.7/0.791 &18.2 &0/0.002/0.4/7.9/37.2/63.5  \\
		\hline
		TinyFaces \cite{hu2017finding}  &ResNet-101 & Box  &56.7/52.9/61.1  &272.4/764.9/0.750  & 59.8 & 4.2/22.6/59.1/\textbf{90.0}/\textbf{93.1}/\textbf{89.6}    \\
		\hline
		VGG+GPR \cite{gao2019c,gao2019domain}  &VGG-16 & Point & 52.5/55.8/49.6  &127.3/439.9/0.410  & 37.4 & 3.1/27.2/49.1/68.7/49.8/26.3 \\
		\hline
		RAZ\_Loc \cite{liu2019recurrent} &VGG-16 & Point &59.8/66.6/54.3 &151.5/634.7/0.305 &42.4 & 5.1/28.2/52.0/79.7/64.3/25.1   \\
		\hline
		Crowd-SDNet \cite{wang2021self} &ResNet-50& point &63.7/65.1/62.4 	&-/-/- &55.1& 7.3/43.7/62.4/75.7/71.2/70.2\\
		\hline
		PDRNet*  &Unknown &Unknown & 65.3/67.5/63.3  & 89.7/\textbf{348.9}/0.261  & 47.0 & 7.4/38.9/63.2/82.9/65.0/24.6 \\
		\hline
		TopoCount \cite{abousamra2020localization} &VGG-16 &Point & 69.2/68.3/70.1  &107.8/438.5/-   & \textbf{63.3} & 5.7/39.1/72.2/85.7/87.3/89.7 \\
		\hline
		RDTM \cite{liang2021reciprocal} &VGG-16 &Point & 69.9/75.1/65.4  &97.3/417.7/0.237   & 45.7 & 11.5/\textbf{46.3}/68.5/74.9/54.6/18.2 \\
		\whline
		IIM  &VGG-16&Point  &73.2/77.9/69.2   &96.1/414.4/0.235 &58.7 & 10.1/44.1/70.7/82.4/83.0/61.4   \\
		\hline
		IIM  &HRNet &Point  &76.0/82.9/70.2   &88.9/412.5/0.161           &49.1 & 11.7/45.3/\textbf{73.4}/83.0/64.5/16.7 \\
		\hline
		IIM  &HRNet&Box  &\textbf{76.2}/81.3/\textbf{71.7} &\textbf{87.1}/406.2/\textbf{0.152} &61.3 & \textbf{12.0}/46.0/73.2/85.5/86.7/64.3   \\
		\whline
	\end{tabular}
	\label{table:nwpu}
\end{table*}

\subsection{Dataset}
\textbf{NWPU-Crowd} \cite{gao2020nwpu} benchmark is the largest and most challenging open-source crowd dataset at present. It contains head points and box labels. There are a total of 5,109 images and 2,133,238 annotated instances.

\noindent\textbf{Shanghai Tech} \cite{zhang2016single} contains two subsets: \textbf{Part A} has 482 images with a total of 241,677 instances, and \textbf{Part B} contains 716 images, including 88,488 labeled heads.

\noindent\textbf{UCF-QNRF} \cite{idrees2018composition} is a dense crowd dataset, consisting of 1,535 images, with a total of 1,251,642 instances.

\noindent\textbf{FDST} \cite{fang2019locality} consists of 100 videos that are captured from 13 different scenes. It contains 15,000 frames, with a total of 394,081 heads, including the point and box labels.

\subsection{Implementation Details}
\setulcolor{red}
\textbf{Training settings. } For the above datasets, the original size image are used. The common data preprocessing schemes, Random horizontally flipping, scaling ($0.8 \times \sim1.2 \times$) and cropping (512$\times$1024) are exploited to augment training data. The batch size is $12$. The learning rates of the confidence predictor and threshold encoder are initialized to $10^{-5}$ and $10^{-6}$, respectively. And the learning rate is updated by a decay rate of 0.99 every epoch. Adam \cite{kingma2014adam} algorithm is used to optimize our framework. The best-performing model on the validation set is selected to conduct testing and evaluate our models.

\textbf{Label Generation.} There are two types of IIM are used in the proposed framework. 1) \textbf{From box annotation}: According to the box-level annotations, overlapping boxes are scaled down until the distance between them is more than 1/4 of the width or height. Finally, the independent instance maps are produced by setting box regions as $1$ and backgrounds as $0$. 2) \textbf{From point annotation}: Similar to the TopoCount \cite{abousamra2020localization}, we dilate the dot mask up to 31 pixels and limit the nearest neighbor to overlap. To evaluate the localization performance, the scale information of heads is required. However, only NWPU-Crowd and FDST datasets provide the bounding box annotation. So, we exploit the box labels provided by NWPU-Crowd to train a head scale prediction network, which automatically generates box-level annotation for the datasets only with point annotation. Please refer to the supplementary for a detailed process.

\subsection{Metrics}

Following the previous crowd localization challenge, NWPU-Crowd \cite{gao2020nwpu}, instance-level Precision (Pre.), Recall (Rec.) and F1-measure (F1-m) are utilized to evaluate models. All results are calculated under large $\sigma$, namely $\sigma_l={\sqrt {{w^2} + {h^2}} }/2 $, where $w$ and $h$ are the width and height of the instance, respectively. At the same time, the proposed model are evaluated on NWPU-Crowd counting task using MAE, MSE and NAE.

\subsection{Ablation Study on NWPU-Crowd}
The ablation study is conducted to explore what factors contribute to the significant performance of our localization approach. For the proposed segmentation-based localization method, we will explore the impact of the fixed thresholds, learnable thresholds (image and pixel-level thresholds), and the working area of L1 loss on the overall framework. Specifically, the following basic experimental settings are described for facilitating understanding.

\textbf{CP}: Confidence Predictor, a fixed global threshold to segment confidence map as an independent instance map.

\textbf{CP+IBM}: Exploit CP and the learnable Image-level Binarization Module (IBM) to locate the crowd.

\textbf{CP+PBM}: Exploit CP and the learnable Pixel-level Binarization Module (PBM) to locate the crowd.

\begin{table}[htbp]	
	\centering
	\caption{Ablation study on different methods and the L1 loss working area. The reported results are tested on the NWPU-Crowd \emph{val set}. The third column indicates whether the L1 loss is used and the gradient range of L1 loss. The backbone is HRNet. }
	\scriptsize
	\setlength{\tabcolsep}{0.95mm}{\begin{tabular}{c|c|cIc|c|cIc|c|c}
			\whline
			\multirow{2}{*}{Method}  & Threshold &\multirow{2}{*}{L1 loss}&\multicolumn{3}{cI}{Localization} &\multicolumn{3}{c}{Counting} \\
			\cline{4-9}
			&type        &  &\textbf{F1-m} & Pre & Rec &MAE &MSE&NAE	\\
			\whline
			CP (Baseline)  &fixed:0.5 & w/o    &74.2 &92.4 & 62.0 &131.4&645.5 &0.237	\\
			\hline
			CP (Baseline)  &fixed:0.8 & w/o    &62.6   &94.1   &46.9 &197.4 &734.8 &0.420 	\\
			\whline
			CP+IBM       &learnable   &IBM     &76.8 &80.6 &73.3&67.9&516.9 &\textbf{0.151}	\\
			\hline
			CP+IBM       &learnable   &CP\&IBM &79.4 &81.8 & \textbf{77.1}	&59.4&449.0 &0.275\\
			\hline
			CP+PBM       &learnable   &PBM     &79.5 &\textbf{86.6}  &73.4 &70.4 &404.6 &0.181\\
			\hline
			CP+PBM (IIM) &learnable   &CP\&PBM  &\textbf{80.2} &84.1 &76.6 &\textbf{55.6}&\textbf{330.9} &0.197	\\
			\whline			
	\end{tabular}}
	\label{Table:ablation}
\end{table}

\textbf{Effect of the IBM/PBM.} \quad
Table \ref{Table:ablation} shows that the method equipped with IBM or PBM has a significant improvement in localization and counting performance compared with the baseline using fixed thresholds. Besides, in the fixed threshold scheme, the localization precision increases with improving the threshold value, but other metrics drop a lot, such as a significant reduction on recall rate and substantial increment on counting error. In the learnable threshold scheme, all metrics are trade-off because the threshold will continuously adapt to the confidence map. Taking the final model (IIM) and the baseline (fixed:0.5) as an example, the F1-measure increased by $8.1\%$ and MAE decreased by $57.7\%$.

\begin{figure*}[t]
	\centering
	\includegraphics[width=0.95\textwidth]{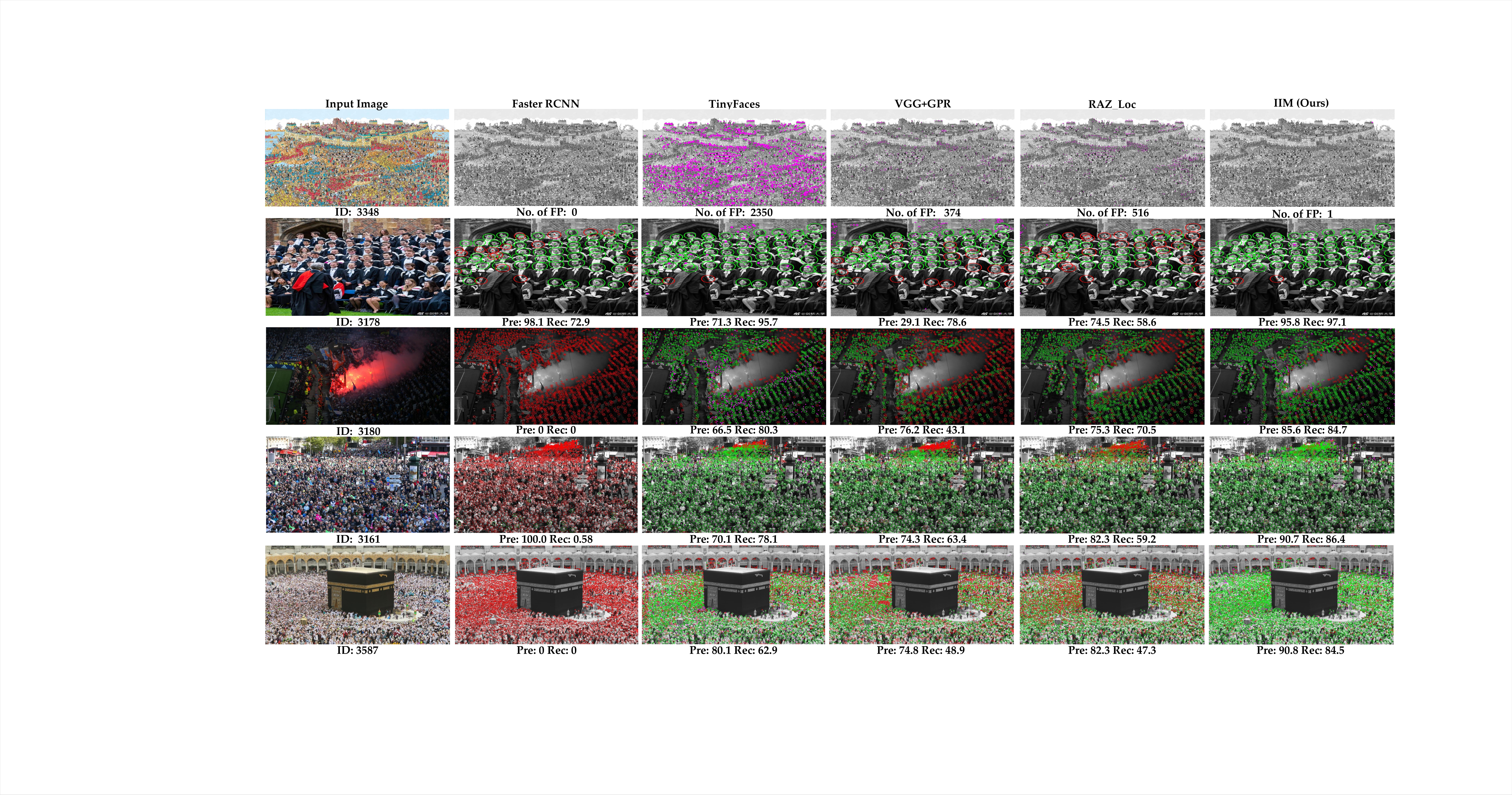}
	\caption{Some typical visualization results of four popular methods and the proposed IIM on NWPU-Crowd \emph{validation set}. The {\color{green}{green}}, {\color{red}{red}} and {\color{magenta}{magenta}} points denote true positive (TP),  false negative (FN) and  false positive (FP), respectively. The {\color{green}{green}} and {\color{red}{red}} circles are GroundTruth with the  radius of $\sigma_l$. For easier reading, images are transformed to gray-scale data.}
	\label{fig:re}
\end{figure*}

\textbf{Effect of the L1 loss.}\quad In the IBM and PBM, the threshold encoder must use the gradient returned by L1 loss for training. But it is optional for the confidence predictor to use L1 loss. Row 4 and 5, Row 6 and 7 in Table \ref{Table:ablation}  show that the IBM and PBM both get better performance by backing the gradient of L1 loss into confidence predictor. It verifies the proposed binarization module can achieve better segmentation and contribute to the feature extraction ability. Overall, the binarization layer brings this effect as it can backpropagation the L1 loss.

\textbf{IBM \emph{vs.} PBM.}\quad This comparison is carried out to look for the best threshold learning solutions among IBM and PBM. In Table \ref{Table:ablation}, Row 4 and 5 show the image-level threshold schemes can obtain well performance. However, Row 6 and 7 show the pixel-level threshold learning scheme can bring more precise localization, contributing to the other metrics' improvement. The pixel-level thresholds are analyzed in the later Section \ref{pbm}.

\subsection{Performance on NWPU-Crowd}

In this section, we compare the proposed IIM with state-of-the-art methods on the NWPU-Crowd benchmark. Table \ref{table:nwpu} lists the Overall performances (Localization: F1-m, Pre. and Rec.; Counting: MAE, MSE and NAE) and per-class Rec. at the box level. By comparing the primary key (Overall F1-m and MAE) for ranking, the proposed IIM achieves first place (F1-m of 76.2\% and MAE of 87.1) in all crowd localization methods. Per-class Rec. on Box Level shows the sensitivity of the model for the instances with different scales. From the table, IIM produces good results in tiny objects (A0$\sim$A2). Besides, it also shows the competitive performance in large objects (A3$\sim$A5). 
In general, IIM balances the localization ability in different scales.

To intuitively demonstrate the localization results, we follow the visualization tools \cite{gao2020nwpu} to compare four traditional methods and the proposed IIM. Fig. \ref{fig:re} illustrates five groups of typical samples (negative sample, sparse and dense crowd scenes ) on NWPU-Crowd \emph{validation set}. For the first sample, Faster R-CNN produces a perfect result, while other methods mistakenly detect some heads. Specifically, the proposed IIM outputs only one head, which is better than TinyFaces, VGG+GPR and RAZ\_Loc. For a sparse scene (Row 2 in Fig. \ref{fig:re}), IIM is the best in all methods, which can detect large-scale heads and yield few false positives. In the dense crowd scenes (Sample 3180, 3161 and 3587),  the general object detection framework, Faster R-CNN misses almost all the heads. TinyFaces, VGG+GPR and RAZ\_Loc perform better but output many false positives and false negatives. The proposed IIM is the best when facing extremely congested scenes. In general, IIM can handle diverse crowd scenes and show the robustness for background objects and negative samples.

\begin{table*}[htbp]
	\centering
	\small
	\caption{The performance of three classical crowd localization methods and the proposed IIM on the four datasets.}
		\begin{tabular}{cIcIcIc|c|cIc|c|cIc|c|cIc|c|c}
			\whline
			\multirow{2}{*}{Method} & \multirow{2}{*}{Backbone}& \multirow{2}{*}{Label} &\multicolumn{3}{cI}{ShanghaiTech Part A} &\multicolumn{3}{cI}{ShanghaiTech Part B} &\multicolumn{3}{cI}{UCF-QNRF} &\multicolumn{3}{c}{FDST}\\
			\cline{4-15}
			&&& \textbf{F1-m} & Pre. &Rec. &\textbf{F1-m} & Pre. &Rec. & \textbf{F1-m} & Pre. &Rec. &\textbf{ F1-m} & Pre. &Rec.\\
			\hline
			TinyFaces \cite{hu2017finding} &ResNet-101& Box&57.3 &43.1 &\textbf{85.5} &71.1 &64.7 &79.0 &49.4 &36.3 &\textbf{77.3} &85.8 &86.1 &85.4  \\
			\hline
			RAZ\_Loc\cite{liu2019recurrent} &VGG-16&Point &69.2 &61.3 &79.5 &68.0  &60.0 &78.3 &53.3 &59.4  &48.3 &83.7 &74.4&95.8  \\
			\hline	
			LSC-CNN \cite{sam2020locate} &VGG-16&Point &68.0&69.6 &66.5 &71.2 &71.7 &70.6 &58.2 &58.6 &57.7 &- &- &-  \\
			\whline
			IIM (ours) &VGG-16 & Point&71.3 &74.0 &68.8 &82.1 &\textbf{92.0} &74.1 &68.4 &73.1 &64.2 &94.1 &\textbf{95.4}&92.9  \\
			IIM (ours) &HRNet  & Point&73.3 &76.3 &70.5 &83.8 &89.8 &78.6 &71.8 &73.7&70.1 &95.4 &\textbf{95.4}&95.3  \\
			\hline
			IIM (ours)  &VGG-16&Box &72.5  &72.6  &72.5   &80.2   & 84.9  &76.0   &68.8   &78.2   &61.5   &93.1  &92.7  &93.5   \\
			\hline
			IIM (ours) &HRNet &Box &\textbf{73.9} &\textbf{79.8} &68.7 &\textbf{86.2} &90.7 &\textbf{82.1} &\textbf{72.0} &\textbf{79.3} &65.9 &\textbf{95.5} &95.3&\textbf{95.8}  \\
			\whline
	\end{tabular}
	\label{table:sota}
\end{table*}

\subsection{Comparison between IIM and other GT}

\begin{figure}[tbp]
	\centering
	\includegraphics[width=0.48\textwidth]{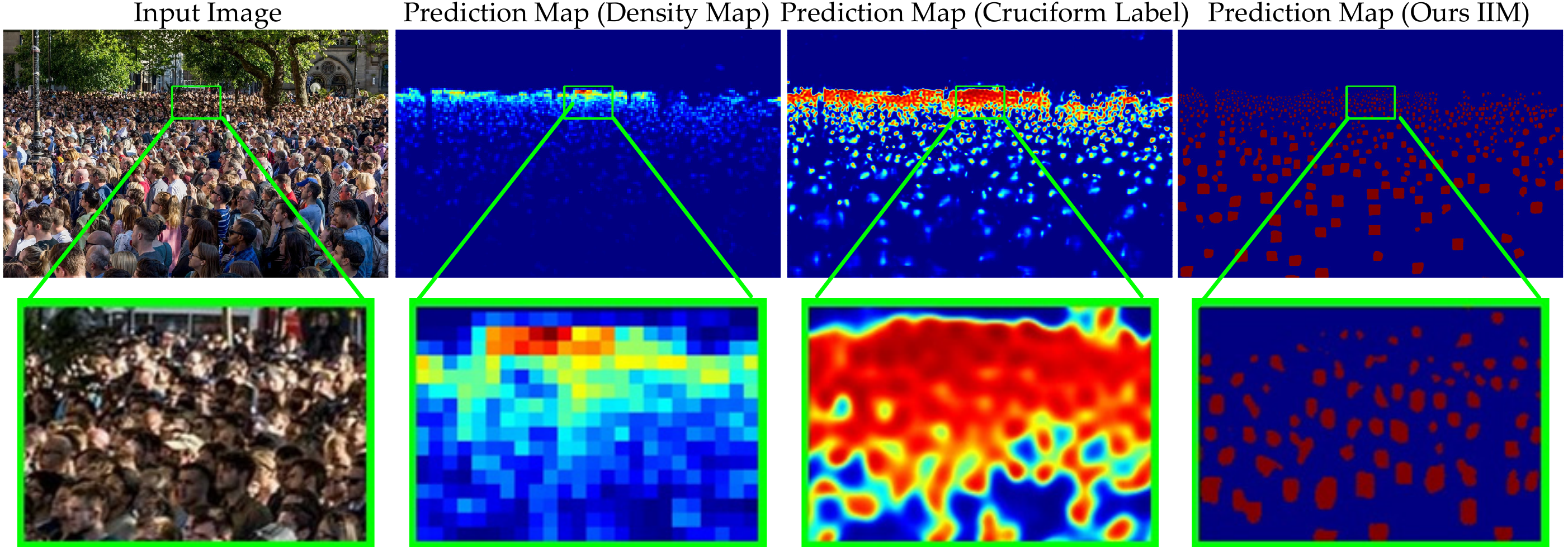}
	\caption{Visual prediction maps for crowd localization  with three types of groundtruth. }
	\label{fig:PredCompare}
\end{figure}

Section \ref{Sec:introduction} compares four different ground truth for the localization task. Here, the prediction results supervised by the different groundtruth are shown to find the best choice for the localization task. As depicted in Fig. \ref{fig:PredCompare}, density map and cruciform label produce coarse prediction results in dense areas. It is very detrimental to subsequent localization. The last column shows the prediction map of the IIM, which shows that the most independent instances are generated under the IIM. This significant improvement makes the localization tasks more effective and straightforward.

To impartially compare three types of GT forms, the corresponding post-processing methods using the same network (VGG-16+FPN) are conducted to obtain the localization results. Table \ref{Table:gt_compare} reports the proposed method achieved better performance in localization and counting. It reveals that the IIM is more suitable for the crowd localization task.

\begin{table}[h]
	\centering
	\caption{The performance on NWPU-Crowd \emph{val set} with three types of groundtruth. They supervises a VGG-16+FPN framework respectively except the post-processing for getting head position. }
	\scriptsize
	\setlength{\tabcolsep}{2mm}{\begin{tabular}{c|c|c|c}
			\whline
			Groundtruth  & Localization &Localization &Counting \\
			type      & Method       &\textbf{F1-m}/Pre/Rec &MAE /MSE/NAE	\\
			\whline
			Density Map  &GPR \cite{gao2019domain}      &57.5/62.4/53.3   &87.4/453.4 /0.830	\\
			\hline
			Cruciform Label   &RAZ\_Loc \cite{liu2019recurrent} &64.2/71.2/58.5 &119.8/626.2 /0.489	\\
			\whline
			IIM (Ours)     &PBM &\textbf{77.0}/\textbf{80.2}/\textbf{74.1} &\textbf{62.7}/\textbf{299.2} /\textbf{0.330}	\\
			\whline			
	\end{tabular}}
	\label{Table:gt_compare}
\end{table}

\subsection{Visual Analysis of the PBM}
\label{pbm}

\begin{figure}[h]
	\centering
	\includegraphics[width=0.45\textwidth]{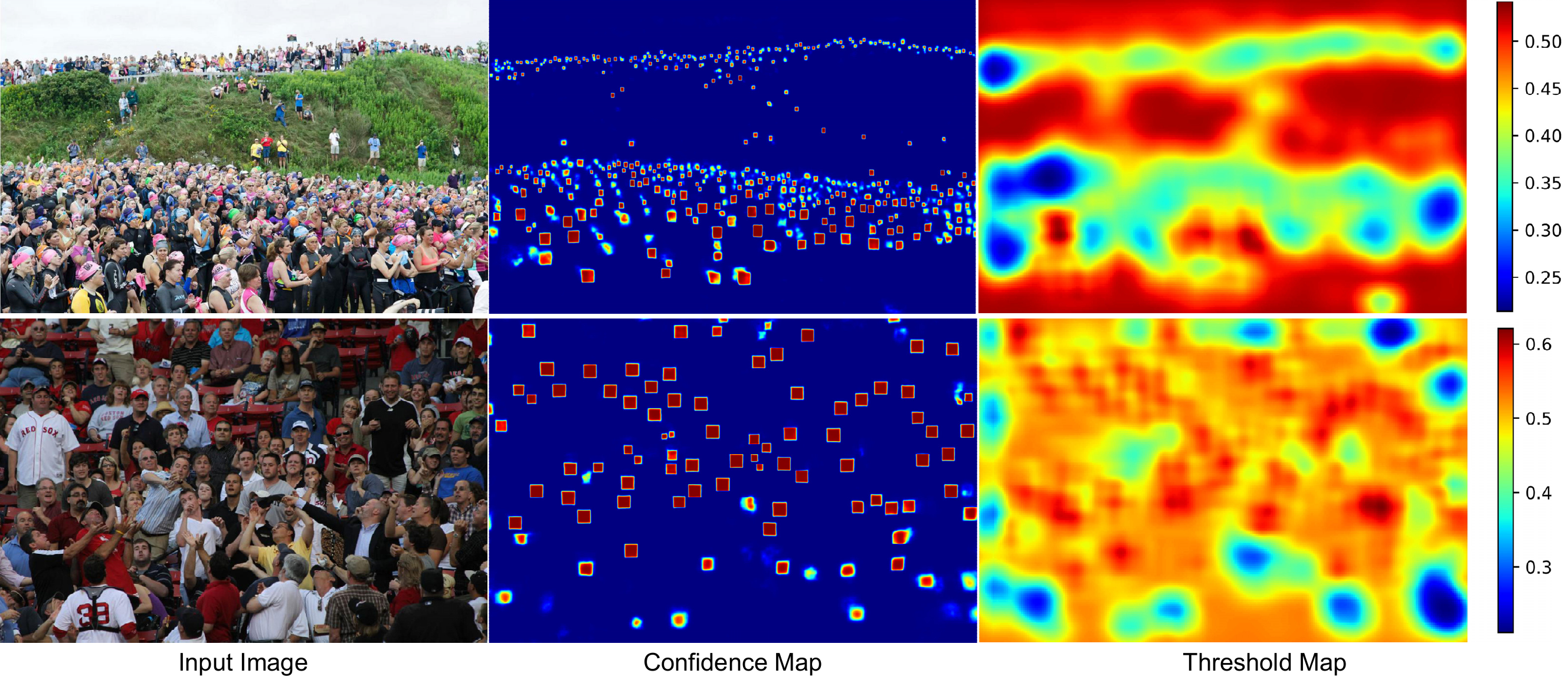}
	\caption{The visualization results of the confidence maps and the pixel-level threshold maps. The backgroug regions are with higher thresholds and the foreground regions are with lower thresholds.}
	\label{fig:PBM}
\end{figure}

PBM is designed to generate threshold adaptively according to different confidence.  Fig. \ref{fig:PBM} shows the thresholds are higher in the background area and lower in the foreground area. It verifies the results obey the Eq. \ref{Eq:7} and Eq. \ref{Eq:8}. Besides, the regions with low confidences (usually large and small heads) also have lower thresholds, which can help improve the localization performance in large and small heads.

\subsection{Robustness on Negative Samples}

\begin{table}[htbp]
	\centering
	\caption{The results (MAE/MSE) of Negative Samples on the NWPU-Crowd \emph{test set}. ``*'' represents anonymous submission.}
	\small
	\begin{tabular}{cccc}
		\whline
		Method	&Main Task & MAE &MSE\\
		\whline
		MCNN \cite{zhang2016single} & Counting & 356.0 & 1232.5 \\
		CSRNet \cite{li2018csrnet} & Counting &176.0&572.3  \\
		S-DCNet \cite{xiong2019open} & Counting &88.5&506.9\\
		AutoScale \cite{xu2019learn} & Counting &123.1&446.7  \\
		DM-Count \cite{wang2020distribution} & Counting &146.7&736.1\\
		\hline
		TinyFaces \cite{hu2017finding} & Localization &52.9 &\textbf{139.2} \\
		RAZ\_Loc \cite{liu2019recurrent} & Localization &60.6 &236.7   \\
		\hline
		PDRNet* & Unknown & 97.2 & 413.4 \\
		\whline
		IIM (VGG-16)  &Localization &31.63 &139.9  \\
		IIM (HRNet)  &Localization &\textbf{31.56} &196.9  \\
		\whline
	\end{tabular}
	\label{Table:NS}
	
\end{table}

Recently, researchers begin to pay attention to the generalization ability and robustness of the crowd model \cite{gao2020nwpu,gao2020cnn}. In this section, we deeply compare the counting performance on the sub-set of NWPU-Crowd, Negative Samples, which consists of some scenes without people and is similar to crowd scenes. Table \ref{Table:NS} reports the results of some popular methods on the NWPU-Crowd's Negative Samples. The ``Main Task'' indicates the task type of the method. From the table, we find the localization methods perform better than the counting method on Negative Samples. For the estimated errors (MAE and MSE), the proposed method achieves the lowest MAE and second-lowest MSE. It shows IIM is very robust to negative samples and backgrounds.
\subsection{Localization Performance on Other Datasets}

Table \ref{table:sota} lists the localization results of three typical open-sourced algorithms: TinyFaces \cite{hu2017finding}\footnote{ https://github.com/varunagrawal/tiny-faces-pytorch}, RAZ\_Loc \cite{liu2019recurrent}\footnote{https://github.com/gjy3035/NWPU-Crowd-Sample-Code-for-Localization} and LSC-CNN \cite{sam2020locate}\footnote{https://github.com/val-iisc/lsc-cnn}. Notably, we implement the first two methods using their default training parameters. For LSC-CNN, we directly test the pre-trained model provided by the authors and evaluate them using our metrics. From the table \ref{table:sota}, the proposed IIM attains the best F1-measure and Precision on the four datasets. In the two dense crowd datasets (ShanghaiTech A and UCF-QNRF), TinyFaces is the best in Rec. By comparing the detailed performance, TinyFaces and RAZ\_loc tend to output higher Recall but lower Precision; LSC-CNN brings the most balanced Recall and Precision. For our method, it produces higher Precision and lower Recall. In general, the proposed method outperforms other SOTA methods. Take F1-m as an example, the relative performance of IIM with HRNet is increased by an average of 15.7\% on these four datasets.

\section{Conclusion}

This paper proposes an effective crowd localization framework, IIM, which outputs independent instance maps to localize each head in crowd scenes. Compared with traditional labels forms (\emph{e.g.} density/dot maps), instance maps are more suitable for crowd localization. Besides, a differentiable Binarization Module is presented to learn a threshold map, which improves the network to yield elaborate instance maps. Moreover, the proposed IIM outperforms other methods in crowd localization. In the future, we will work on localizing extremely small heads in the dense crowds.

{\small
\bibliographystyle{IEEEtran}
\bibliography{IEEEabrv,reference}
}

\end{document}